\documentclass[journal]{IEEEtran}
\ifCLASSINFOpdf
\else
\fi

\usepackage{graphicx}
\usepackage{subfigure}
\usepackage[linesnumbered,boxed,ruled,commentsnumbered]{algorithm2e}
\begin{document}

\title{A simple and effective postprocessing method for image classification}
%
%
%

\author{Yan~Liu,~\IEEEmembership{}
        Yun~Li*,~\IEEEmembership{}
        Yunhao~Yuan,~\IEEEmembership{}
        and~jipeng~Qiang,~\IEEEmembership{}
\thanks{Yan Liu, Yun Li, Yunhao Yuan and jipeng qiang are with the Department of School of Information Engineering, Yangzhou University, Yangzhou, China \textit{(Corresponding authors: Yun Li \{liyun\}@yzu.edu.cn)}}
\thanks{}
\thanks{}}

%
%

\markboth{Journal of \LaTeX\ Class Files,~Vol.~14, No.~8, August~2015}%
{Shell \MakeLowercase{\textit{et al.}}: Bare Demo of IEEEtran.cls for IEEE Journals}
%



\maketitle

\begin{abstract}
{W}{ether} it is computer vision, natural language processing or speech recognition, the essence of these applications is to obtain powerful feature representations that make downstream applications completion more efficient. Taking image recognition as an example, whether it is hand-crafted low-level feature representation or feature representation extracted by a convolutional neural networks(CNNs), the goal is to extract features that better represent image features, thereby improving classification accuracy. However, we observed that image feature representations share a large common vector and a few top dominating directions. To address this problems, we propose a simple but effective postprocessing method to render off-the-shelf feature representations even stronger by eliminating the common mean vector from off-the-shelf feature representations. The postprocessing is empirically validated on a variety of datasets and feature extraction methods.such as VGG, LBP, and HOG. Some experiments show that the features that have been post-processed by postprocessing algorithm can get better results than original ones.
\end{abstract}


%
\IEEEpeerreviewmaketitle

\section{Introduction}
%
%
%
%
\IEEEPARstart{W}{ether} it is computer vision\cite{tian2018deep}, natural language processing\cite{deng2018deep} or speech recognition\cite{swain2018databases}, the essence of these applications is to obtain powerful feature representations that make application completion more efficient. Image recognition has always been a research hotspot for many researchers. In addition, Large Scale Visual Recognition Challenge (ILSVRC)\footnote{http://www.image-net.org/challenges/LSVRC/} also attract the attention of many researchers. In the early stage of the study on image recognition, some hand-craft features are employed, such as LBP\cite{ojala1996comparative}, HOG\cite{lowe1999object} and SIFT\cite{lowe2004distinctive}. However, these features tend to be very high in dimension and have a lot of redundant information, which often leads to dimensional disasters. To address the above mentioned problems, some subspace learning methods perform dimension reduction on these hand-craft features and then substituting them into appropriate classifiers, such as PCA\cite{jolliffe1986principal}, LDA\cite{belhumeur1997eigenfaces} and CCA\cite{hotelling1936relations}. Benefit from subspace learning methods, on some small-scale issues, these hand-craft features can fully understand the feature representation of the image, so that these methods can meet the basic application needs. 

In recent years, due to the increasing computer computing power and image acquisition methods have been continuously enhanced, we have witnessed the tremendous revolution brought by deep neural networks in computer vision. Various convolutional neural network methods constantly refresh the rankings of major tasks. For the first time, krizhevsky et al\cite{krizhevsky2012imagenet}. used the relu activation function in a convolutional neural network, which brought a huge success for image recognition tasks. ZFNet\cite{zeiler2014visualizing} improved krizhevsky's method by first reducing the size of the convolution kernel in the first layer from 11x11 to 7x7, while reducing the step size during convolution from 4 to 2. This expands the intermediate convolutional layer to capture more information. VGGNet\cite{simonyan2014very} extends the depth of the network to 19 layers and uses a small size convolution kernel of 3x3 at each convolution level. The results prove that depth has an important impact on network performance. GoogleNet\cite{szegedy2015going} also increases the breadth and depth of the network, and significantly improves network performance without a significant amount of computation compared to a narrower and shallower network. In order to better meet the task of image recognition, changes based on various convolutional neural network frameworks have been proposed, making the feature representation more powerful. The feature representation is mainly divided into two aspects to improve. The first is to remove the redundant information of the image features, such as Dropout\cite{hinton2012improving} by throwing away some neurons randomly, essentially, throwing away some redundant information represented by image features, thus to some extent against overfitting. Corresponding to dropout\cite{ghiasi2018dropblock}, dropblock randomly throws away some feature maps in the convolutional layer so that redundant pixel features are thrown away. the second aspect is to make the features more discriminative by some mapping methods. such as SS-HCNN\cite{chen2018ss} identifies image hierarchy using a newly designedlarge-scale MMC technique, and groups images into different visually compact clusters at different hierarchical levels, which make the feature representation has the ability of discrimination.

Different from the above feature enhancement method, in this paper, we find that a simple and effective processing method renders the off-the-shelf existing representations even stronger. The proposed algorithm is motivated by the NLP application\cite{mu2018all} and the following observation.

\textbf{observation:} all image feature representations we tested, have the following observation. 
\begin{itemize}
\item[1)] the image feature representations have non-zero mean, which indicates feature vectors share a large common vector.
\item[2)] If the common mean vector is removed, the representations are far from isotropic. indeed, much of the energy of most feature representation is stacked in a very low dimensional subspace.
\end{itemize}
To our best knowledge, the same common vector and the same dominating directions have side effects on the feature representation. In this paper, we propose a feature postprocessing algorithm to eliminate them in two steps.
\begin{itemize}
\item[step1:]effectively reducing the energy by removing the nonzero mean vector from all image feature vectors
\item[step2:]projecting away from the dominating $T$ directions, reducing the dimension effectly.
\end{itemize}
Regarding the value of $T$, we will discuss in the experimental section.

Finally, we emphasize that our feature postprocessing algorithm is counter intuitive - particularly traditional denoising  is to eliminating the weakest direction (in a singular value decomposition of the stacked image feature vectors) instead of the dominating direction. 

The rest of the papaer is organized as follows. we give a brief background on our postprocessing algorithm. next, we will analyze the differences between human visual methods and computer vision, and then lead to our postprocessing algorithm. In addition, we will give a possible explanation mathematically in section 4. Experiment results are provided in section 5. Finally, we give a conclusion in section 6.
\section{related work}
In the previous study, the idea of removing the top principal components has been preliminary studied. Price et al.\cite{price2006principal} firstly proposed the idea and applied it to population matrix analysis. In addition, in NLP applications, Bullinaria et al\cite{bullinaria2012extracting} assume that the maximum variance component of the symbiosis matrix is destroyed by information other than lexical semantics,  the removal of the top principal components was proved to be reasonable. 
\section{The visual models of Human and Computer}
For humans, human beings distinguish an object, usually by finding the difference with other objects. then those difference determine which object this object belongs to. The process of finding difference usually involves three steps, The first step is to see the whole object, the second step is to find the similarity of each object, and finally the whole object minus this similarity is the special point of the object itself, This special point is the key to distinguishing different objects by human beings. However, when the computer imitates human visual logic, it always stays in the first step, considering only all the features of an object, so that it logically brings a lot of similar features to the classification of the final object. Obviously, this part of the information is highly "redundant". In this section, we will explain mathematically the visual logic of humans and computers, then provide a seemingly viable computer vision logic.
\begin{figure} \centering 
\subfigure[cat1] { \label{fig:a} 
\includegraphics[width=0.4\columnwidth]{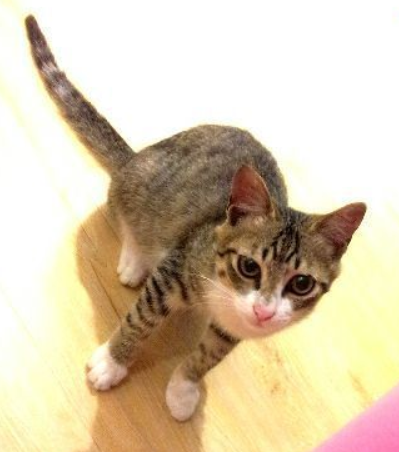} 
} 
\subfigure[cat2.] { \label{fig:b} 
\includegraphics[width=0.4\columnwidth]{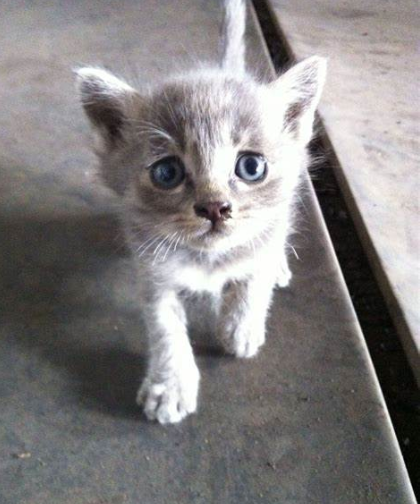} 
} 
\caption{Two different cats} 
\label{fig} 
\end{figure} 
\subsection{the visual model of human}
 Fom the Fig \ref{fig}, we can easily distinguish the different cat. Now, we simulate the process of distinguish cats in our brains. The first step is to create the  “full impression” of two cats in our brains. The second step is to find the difference between the two cats and similarities. In the last step, we use similarities and differences to distinguish whether the two cats are a type of cat. if give this process a mathematical expression, we can get:
\begin{equation}
{I_{full{\rm{ }}impression}}\approx {S_{similarities}} + {D_{difference}}
\end{equation}
where ${I_{full{\rm{ }}impression}}$ indicates the overall image of an object in the human brain, ${S_{similarities}}$ indicates the similarity between this object and other objects that has been seen and ${D_{difference}}$ indicates the difference with those. Then the recognition mechanism of the human brain can distinguish this object is most likely to belong to which object through ${S_{similarities}}$ and ${D_{difference}}$, however, this process of recognition mechanism is hard to simulate completely. The current exiting visual model of computer strategy can only be imitated simply in computer vision, next, we will give a further discussion about the computer vision model from the perspective of human visual logic.
\subsection{the visual model of computer}
In pattern recognition,  Regardless of the features extracted manually, or features extracted by convolutional neural networks, the entire feature of the object is extracted as much as possible. It may be effective for some tasks, such as image caption\cite{xu2015show}, but for pattern recognition, this approach seems to bring huge redundant information to the final classifier. from the visual model of human, we can see only ${D_{difference}}$ is good for distinction and ${S_{similarities}}$ is almost useless for distinction. Because human beings are inherently intelligent, it is easy to discard unwanted information, however it is difficult for computer to discard. For further discussion, we give the visual model of computer mathematical expression:
\begin{equation}
f{c_{entirefeature}} \approx S{c_{common}} + D{c_{difference}}
\end{equation}
where $f{c_{entirefeature}}$, $S{c_{common}}$ and $D{c_{difference}}$ respectively correspond to ${I_{full{\rm{ }}impression}}$, ${S_{similarities}}$ and ${D_{difference}}$, $f{c_{entirefeature}}$ indicates the entire feature representations extracted by the feature extraction algorithm, it also can be seen as a pre-classifier feature. $S{c_{common}}$ indicates the shared common vector and $D{c_{difference}}$ indicates the feature representations that distinguishes strongly. In addition, from Equation 2, we can get:
\begin{equation}
D{c_{difference}}\approx f{c_{entirefeature}}-S{c_{common}}
\end{equation}
To our best knowledge, for pattern recognition, the powerful $D{c_{difference}}$ is good for recognition and $S{c_{common}}$ is the redundant information for recognition. our postprocessing's goal is to eliminate the redundant information (shared common vector) from the pre-classifier feature.
\section{postprocessing}
we test our observstions on various feature representations: these feature representations are derived from different data sets by several feature learning algorithms. f(sift)  are extracted from the lfw\cite{huang2008labeled} by sift\cite{dalal2005histograms}, f(Vgg19) trained on cifar10\cite{krizhevsky2009learning} using Vgg19, f(ResNet) trained on cifar10 using ResNet, f(LBP) are extracted from the AT$\&$T dataset\footnote{https://www.cl.cam.ac.uk/research/dtg/attarchive/facedatabase.html} using the lbp operator\cite{ojala1996comparative}, f(Center) trained on lfw\cite{huang2008labeled} using Center loss\cite{wen2016discriminative}. these methods and data sets are the most commonly used for image classification.

\begin{table}
\caption{A detailed description for some feature representations}
\centering
\begin{tabular}{p{38pt}p{38pt}p{38pt}p{38pt}p{38pt}}\label{table1}
	Method:&Image set&dim&image size&${\left\| u \right\|_2}$\\
    \hline
    f(sift)&lfw&1280&13k&5.99\\
     f(VGG19)&cifar10&4096&60k&3.12\\
     f(ResNet)&cifar10&1024&60k&4.05\\
     f(LBP)&AT\&T database&&0.4k&0.0487\\
     f(Center)&lfw&1024&13k& 6.0146\\
     \hline     
\end{tabular}
\end{table}

From table \ref{table1}, we can see that all feature representations we tested have non-zero mean, which indicate our observations exist.

Now, we let $f(i) \in {R^d}$ be a feature representation for a given image $i$ in the dataset A, the number of images in A is $N$, $d$ is the dimension of fearure representation. we can esasily observe there are two phenomenas in each of feature representations listed above:
\begin{enumerate}[(1)]
\item $ \{f(i):i \in A\} $ are not of zero-mean: all $\{f(i)$ share a non-zero common vector, it corresponds to $S{c_{common}}$ in section B.III. $ f(i) = \widetilde {f(i)} + u$, where $u$ is the average of all $\{f(i)$'s. That is to say $
u = {1 \mathord{\left/
 {\vphantom {1 N}} \right.
 \kern-\nulldelimiterspace} N}\sum\limits_{i = 1}^N {f(i)}$. From the table1, the norm is about 1/6 to 1/2 of the average norm of all presentations.
\item ${\rm{\{ }}\widetilde {f(i)}:{\rm{ }}i \in A\}$ are not isptropic: we get the principal component of ${\rm{\{ }}\widetilde {f(i)}:{\rm{ }}i \in A\}$ by principal component analysis, in this paper, let ${u_1},...,{u_d}$ be the first to the last components. In addition, each $\widetilde {f(i)}$ can be linearly combined by $u_i$:$\widetilde {f(i)} = \sum\nolimits_{j = 1}^d {{\alpha _j}(i)} {u_j}$.
\end{enumerate}
\subsection{Algorithm}
Since almost feature representations without postprocessing share the same common vector $u$ and the same dominating directions. such shared vector and directions maybe bring huge redundant information to feature representations, it is not good for pattern classification. Here we propose a preliminary method to eliminate them, as formally achieved as Algorithm 1. 
\IncMargin{1em} 
\begin{algorithm}

    \SetAlgoNoLine 
    \SetKwInOut{Input}{\textbf{Input}}\SetKwInOut{Output}{\textbf{Output}} 

    \Input{
        \\
        The feature representations $ \{f(i):i \in A\} $\;\\
        The feature dimension saved by PCA $d$\;\\
        A threshold parameter $T$\\}
    \Output{
        \\
        Processed representations ${{f^{'}}(i)}$.\\}  
    \BlankLine
    Computer the mean of $\{f(i):i \in A\}$, $\leftarrow \frac{1}{N}\sum\nolimits_{i \in A} {f(i)}$\, \; 
    demeaned value: $\widetilde {f(i)} \leftarrow f(i) - u$\;
    Computer the PCA components: ${u_1},...,{u_d} \leftarrow PCA(\widetilde {f(i)},d)$\;
    process the feature representations: ${f^{'}}(i) \leftarrow \widetilde {f(i)} - \sum\nolimits_{j = 1}^d {\left( {u_j^Tu(i)} \right)} {u_j}$
    \caption{Postprocessing algorithm on feature representation.\label{al3}}
\end{algorithm}
\DecMargin{1em}
\begin{figure*}{}\centering
\includegraphics[width=2\columnwidth]{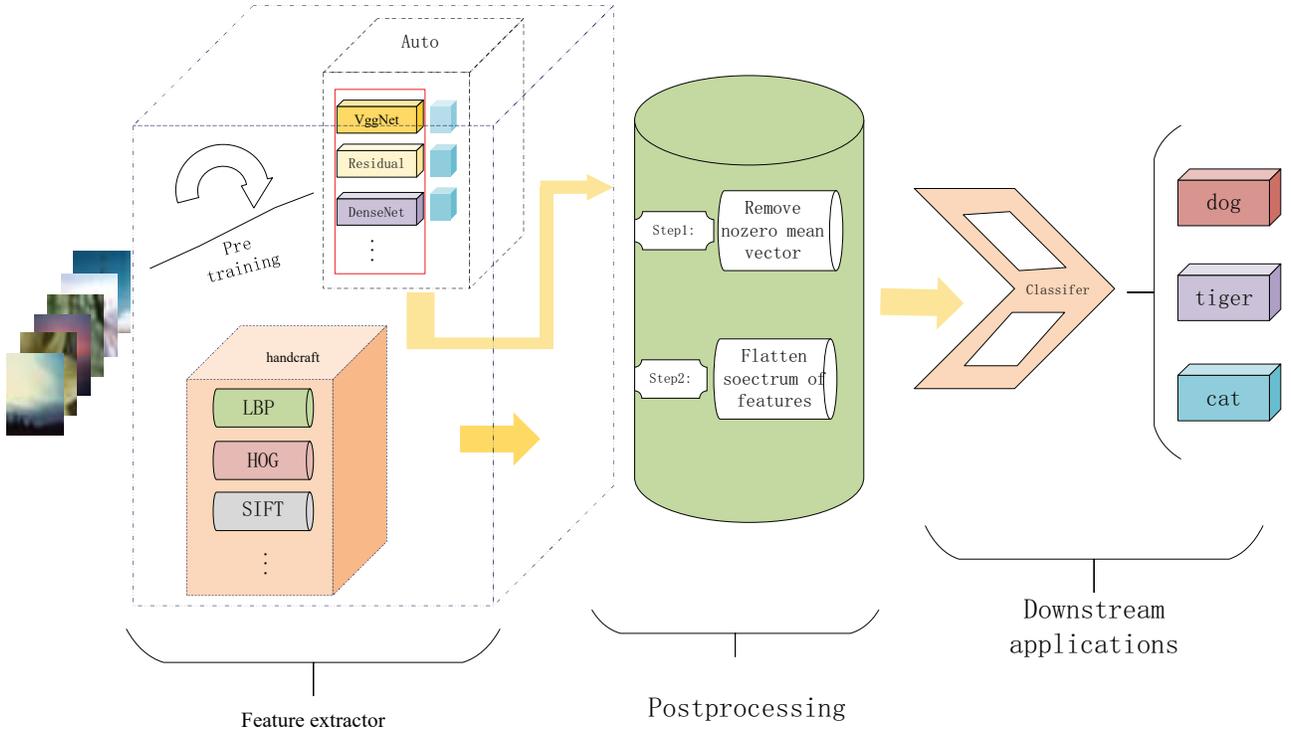} 
\caption{The feature postprocessing algorithm framework we proposed, our feature postprocessing algorithm only cares about the intermediate links of feature extraction and downstream application.} \label{figure2}
\end{figure*}

As shown in Fig \ref{figure2}, it is our framework we proposed. Feature extracted by the feature extractor are purified by our proposed postprocessing algorithm, then it will be applied in downstream applications ( in this paper, we onoly consider the image classification).
\subsection{postprocessing as a "rounding" wowards istropy}
The idea ofisotropy is mainly inspired by the partition function proposed in \cite{arora2016a}.
\begin{equation}
{\rm H}(\omega) = \sum\limits_{i \in A} {\exp ({\omega ^T}f(i))}
\end{equation}
where A represents the entire sample domain, $f(i)$ represents the $i$-th image features. Paper \cite{arora2016a} also gives a proof that ${\rm H}(i)$ should be approximately be a constant with any unit vector $\omega$. In this paper, we still assume that this rule is applicable to image features. Hence, we redefine the measure of isotropy mathematically as follows:
\begin{equation}
{\rm M}(\{ f(i)\} ) = \frac{{\min {\rm{ }}H(\omega )}}{{\max {\rm{ H(}}\omega {\rm{)}}}}{\rm{, }}\left\| \omega  \right\| = 1
\end{equation}
where ${{\rm M}(\{ f(i)\} )}$ belongs to the 0-1 interval, and if ${{\rm M}(\{ f(i)\} )}$ closer to 1 indicates the feature (${f(i)}$) is more isotropic. As far as we known, the isotropy has a “purification” effect on feayure representations. Essentially, our proposed feature postprocessing algorithm is to make ${{\rm M}(\{ f(i)\} )}$ tend to 1 so that features become more isotropic.

Now, let $A$ be the matrix stacked by all image feature vector, where the rows correspond to one image feature vectors, and $I$ be the $|V|$-dimensional vectors with all entries equal to one, then, ${\rm H}(\omega)$ can be obtained by Taylor's Formula as follows:
\begin{equation}
H(\omega ) = \left| A \right| + {I^T}A\omega  + \frac{1}{2}{\omega ^T}{A^T}A\omega  + \sum\limits_{k = 3}^\infty  {\frac{1}{{k!}}} \sum\limits_{i \in {\rm A}} {({\omega ^T}} f(i){)^k}
\end{equation}
Then, we ${{\rm M}(\{ f(i)\} )}$ can be very coarsely estimated as by first order approximation and second order approximation. The specific process is as follows:
\subsubsection*{\textbf{first order approximation}}
\begin{equation}
{\rm M}(\{ f(i)\}  \approx \frac{{\left| A \right| + \min ({I^T}A\omega )}}{{\left| A \right| + \max ({I^T}A\omega )}}
\end{equation}
and if we set $\omega$ to 1, then, we can get:
\begin{equation}
{\rm M}(\{ f(i)\}  \approx \frac{{\left| A \right|{\rm{ - }}\left\| {{I^T}A} \right\|}}{{\left| A \right| + \left\| {{I^T}A} \right\|}}\label{huahua}
\end{equation}
From formula \ref{huahua}, if we let ${{\rm M}(\{ f(i)\} )}$ $=$ 1, then we need to make ${{I^T}A}$ $=$ 0, which is equivalent to make $\sum\limits_{i \in A} {\exp ({\omega ^T}f(i))}$ $=$ 0. It is easily to known the goal of the first order approximation matches with the goal of our first step of our postprocessing algorithm, where we enforce $f(i)$ to have a zero mean.
\subsubsection*{\textbf{Second order approximation}}
\begin{equation}
{\rm M}(\{ f(i)\}  \approx \frac{{\left| A \right|{\rm{ + }}\min ({I^T}A\omega ) + \min (\frac{1}{2}{\omega ^T}{A^T}A\omega )}}{{\left| A \right|{\rm{ + }}\max ({I^T}A\omega ) + \max (\frac{1}{2}{\omega ^T}{A^T}A\omega )}}
\end{equation}
and if we set $\omega$ to 1, then, we can get:
\begin{equation}
{\rm M}(\{ f(i)\}  \approx \frac{{\left| A \right|{\rm{ - }}\left\| {{I^T}A} \right\|{\rm{ + }}\frac{1}{2}\sigma _{\min }^2}}{{\left| A \right| + \left\| {{I^T}A} \right\| + \frac{1}{2}\sigma _{\max }^2}}\label{huahua1}
\end{equation}
where $\sigma _{\max }$ and $\sigma _{\min }$ represent the largest and smallest singular value of $A$. if we let ${{\rm M}(\{ f(i)\} )}$ $=$ 1, then we need to make ${{I^T}A}$ $=$ 0 and $\sigma _{\max }=\sigma _{\min }$ respectively. Actually, the face that $\sigma _{\max }=\sigma _{\min }$ is equal to making the spectrum of $f(i)$ be flat. It is easily to known the goal of the second order approximation matches with the goal of our second step of our postprocessing algorithm, where we remove the highest singular values.

\subsubsection*{\textbf{Empirical Verification}} Since it is impossible to find closed-form solution of $
\arg {\max _{\left\| \omega  \right\|}}H(\omega )$ and $
\arg {\min _{\left\| \omega  \right\|}}H(\omega )$. In addition it is impossible to enumerate all value of $\omega$, we only estimate the measure by:
\begin{equation}
{\rm M}(\{ f(i)\}  \approx \frac{{{{\min }_{\omega  \in U}}H(\omega )}}{{{{\max }_{\omega  \in U}}H(\omega )}}
\end{equation}
where $U$ is the set of eigenvectors of $(A^T)A$. Need to point out that we validate the effect of postprocessing of on ${{\rm M}(\{ f(i)\} )}$ empirically and experimentally. However, we are not giving a completely correct explanation yet. 

\section{experiment}
In this section, we choose several representative methods on cifar10, cifar100, lfw, Caltech-UCSD Birds-200-2011\cite{wah2011caltech} and Stanford cars\cite{KrauseStarkDengFei-Fei_3DRR2013}, those methods contain classic handcrafted features. in addition, now some popular deep learning feature learning methods are also included. These methods are not necessarily the best results on those data set, but they are the most representative methods of those data set. our proposed method is a postprocessing based on these methods named $S$-post, where $S$ is an existing method. Some specific experimental steps will be mentioned in the subsections respectively. 
\subsection{Discussion on parameters T and d}
Here, we further discuss the threshold parameter $T$ and the number of dimensions that the PCA algorithm needs to save $d$ based on the f(DFLA) feature. Since the feature dimension of f(DFLA) feature is 1024 and $d$ is at most 1024. In this work, d from 0 to 1024 in increments of 100. In addition, according to our experience, since $T$ represents the main component information of the throw away feature, the value of $T$ generally ranges[0, 10].

\begin{figure}{}\centering
\includegraphics[width=0.8\columnwidth]{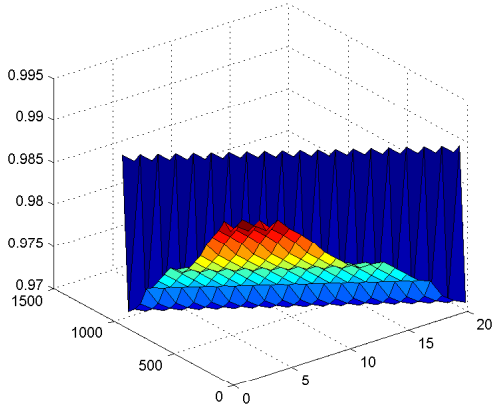} 
\caption{Postprocessing on the f(DFLA) feature, the recognition rate changes with the parameters $T$ and $d$} \label{resultT}
\end{figure}

From the Fig. \ref{resultT}, we can find a very strange phenomenon that $d$ seems to have no effect on the final result, which seems to indicate that the dominating directions are in a low dimensional space. when $T$ is 1, the postprocessing feature is the most powerful. For our best knowledge that $T$ depends on the feature representations generally(In general, different feature extraction methods extract different features, which leads to different dimension of dominating directions that need to be thrown away). In next work, we set $d$ to the dimension of the original feature, and the value of $T$ ranges 1-10, for classification tasks, it is easier to get the powerful feature representations.
\subsection{Face verification}
LFW (Labeled Faces in the Wild) is a data set compiled by the Computer Vision Laboratory of Massachusetts State University Amherst. It is mainly used to study face recognition in unconstrained situations. The LFW database mainly collects images from the Internet, not the lab. It contains more than 13,000 face images. Each image is identified by the corresponding person's name. Among them, 1680 people correspond to more than one image, that is, about 1680. An individual contains more than two faces.
\begin{figure}{}\centering
\includegraphics[width=0.8\columnwidth]{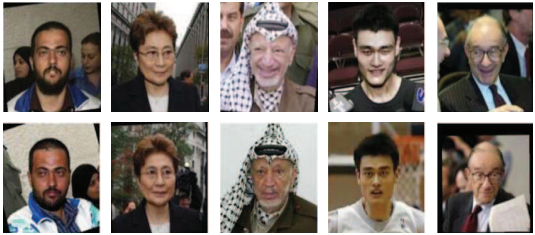} 
\caption{Paired face images in lfw, all the face is funneled} 
\end{figure}

In this experiment, a random selection of 6000 pairs of faces constitutes a face recognition picture pair, 3000 pairs belong to the same person, 3000 pairs belong to different people. The test process LFW gives a pair of photos asking if the two photos in the system under test are the same person, and the system gives the answer of "yes" or "no". The face recognition accuracy rate can be obtained by the ratio of the system answer and the real answer of the 6000 face test results. In addition, some representative methods for lfw are choosen for face vertification, such as Fisher vector face\cite{simonyan2013fisher}, Light CNN\cite{wu2018light}, Center Face(ResNet)\cite{wen2016discriminative}, normface\cite{wang2017normface} and DeepID\cite{sun2014deep}.

\textbf{Fisher vector face}: Fisher vector face are obtained by blending fisher features\cite{belhumeur1997eigenfaces} and sift\cite{lowe2004distinctive}

\textbf{Light Cnn}: Maxout is used in light CNN, it is used as an activation function to achieve noise filtering and retention of useful signals in light cnn, which bring better feature maps named MFM (Max-Feature-Map).

\textbf{Center Face}: Simultaneously learning a center for deep features of each class and penalizing the distances between the deep features and their corresponding class centers, it can get a discriminative feature representation for face recognition.

\textbf{Normface}: It optimizes cosine similarity rather than inner product and introduces an agent vector for each class to rereformulate metric learning.

\textbf{DeepID}: DeepID learns effective high-level features revealing identities for face verification. The features are built on top of the feature extraction hierarchy of CNNs and are fused with multi-scale mid feature representations.
\subsection*{Implementation Details}
For original light cnn and our light cnn-post, we choose the light cnn-9 as the base model, some details about it can be found in \cite{wu2018light}. CASIA-WebFace\cite{yi2014learning} is employed to train the light cnn and all the face images are coverted to gray images and normalized to 144x144. According to the 5 facial points, we not only rotate two eye points horizontally but also set the distance between the midpoint of eyes and the midpoint of mouth $(ec_mc_y)$, and the y axis of midpoint of eyes $(ec_y)$. During training, the initial learning rate is set to $1 \times {10^{{\rm{ - }}3}}$ and gradually decaying to $5 \times {10^{{\rm{ - }}5}}$. For framework of center face and Corresponding center face-post, we use the web face images as training set, such as CACD2000\cite{chen2015face}, Celebrity+\cite{liu2015deep} and CASIA-WebFace\cite{yi2014learning}. finally about 0.7M images of 17,189 unique persons can be gotten. For better results, we use the optimal values discussed in document \cite{wen2016discriminative} for parameters $\lambda$ and $\alpha$, $\lambda$ is 0.6 and $\alpha$ is 0.5. Other experimental details can be seen in \cite{wen2016discriminative} and then we also drawing ydwen's code\footnote{https://github.com/ydwen/caffe-face}. For normface and normface-post, we strictly follow all the experimental settings as \cite{wang2017normface} except replace the final inner-product layer and softmax layer with scaled$\_$cosine$\_$softmax. For DeepID\cite{yi2014learning} and corresponding DeepID-post, we use hqli's code\footnote{https://github.com/hqli/face$\_$recognition} to learning the DeepID face images. From the above introduction, we almost always choose the best method mentioned in original papers to learn face features, which makes it easier to compare the effectiveness of feature postprocessing methods.
\begin{table}
\caption{Face Verification Accuracy of Fisher, Fisher-post, Light Cnn, Light Cnn-post, Center Face, Center Face-post, Normface, Normface-post, DeepID, DeepID-post on lfw}\label{table2}
\centering
\begin{tabular}{lll}
\hline
Methods & Accuracy[$\%$]\\
\hline
Fisher vector face &87.3\\
\textbf{Fisher vector face-post} &\textbf{89.1}\\
Light Cnn &98.13\\
\textbf{Light Cnn-post} &\textbf{98.32}\\
Center Face &99.03 \\
\textbf{Center Face-post} & \textbf{99.15}\\
Normface & 99.11\\
\textbf{Normface-post} & \textbf{99.09}\\
DeepID & 97.33\\
\textbf{DeepID-post} & \textbf{97.45}\\
\hline
\end{tabular}
\end{table}
\subsection*{results}
The Light CNN\cite{wu2018light}, Center Face(ResNet)\cite{wen2016discriminative}, normface\cite{wang2017normface} and DeepID\cite{sun2014deep} 's results of face verification on LFW datasets are present in table \ref{table2}.
From the results in table \ref{table2}, we can see our proposed methods can get better results than original methods in most circumstances, which shows when the common mean is subtracted from the original feature, the new feature becomes more discriminant, which may be beneficial to image classification.

\subsection{Image classification}
In these section, we use the most common image classification datasets cifar10 and cifar100. 
\subsection*{cifar10}
Cifar10 has a total of 60,000 color images, which are 32*32, divided into 10 categories, each with 6000 images. There are 50,000 pieces for training, which constitutes 5 training batches, each batch of 10,000 images; another 10000 is used for testing and constitutes a batch. The data of the test batch is taken from each of the 10 categories, and each type is randomly taken from 1000 sheets. The remaining ones are randomly arranged to form a training batch. Note that the various types of images in a training batch are not necessarily the same number. In general, the training batches have 5000 maps for each category.
\subsection*{cifar100}
Like CIFAR-10, except that cifar100 has 100 classes, each containing 600 images. Each class has 500 training images and 100 test images. The 100 classes in cifar100 are divided into 20 superclasses. Each image has a "fine" label (the class it belongs to) and a "rough" label (the superclass it belongs to).
\subsection*{Implementation Details}
For better results, during training, the ImageDataGenerator function in keras\footnote{https://github.com/keras-team/keras} is employed to do data augmentation for cifar10 and cifar100. The horizontal$\_$flip in imagedatagenerator is set to true, in addition, width$\_$shift$\_$range is set to 0.125, height$\_$shift$\_$range is set to 0.125, fill$\_$mode is set to 'constant', cval is set to 0.
\begin{figure}{}\centering
\includegraphics[width=1\columnwidth]{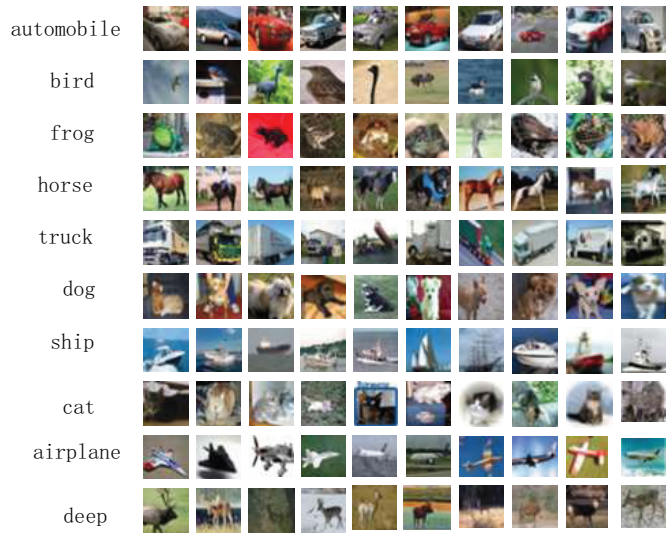} 
\caption{Ten images of ten projects in the cifar10 dataset} 
\end{figure}
\textbf{HOG}: HoG\cite{lowe1999object} descriptor with cell size 4x4 was used to extract features from images. Next, linear SVM\cite{chang2011libsvm} was applied. 

\textbf{LeNet}: LeNet\cite{lecun1998gradient} is the first CNN network proposed by LeCun. During training, batch$\_$size is set to 128, epochs is 200, iterations is 391, weight$\_$decay is set to 0.0001. When the epoch is less than 100, the learning rate is 0.01, when the epoch is morn than 200, the learning rate is 0.01. Relu is used as the activation function, he$\_$normal\cite{he2015delving} ia employed in parameter initialization.

\textbf{VGGNet}: VGGNet\cite{simonyan2014very} is a convolutional neural network developed by researchers at the Visual Geometry Group and Google DeepMind. VGGNet explored the relationship between the depth of convolutional neural networks and their performance. VGGNet successfully constructed 16 to 19 deep convolutional neural networks by repeatedly using 3x3 small convolution kernels and 2x2 maximum pooling layers. Here. we choose Vgg-19 as our base model. During training, batch$\_$size is set to 128, epochs is 200, iterations is 391, weight$\_$decay is set to{} 0.0001. when the epoch is less than 80, the learning rate is 0.1, when the epoch is between 80 and 160, the learning rate is 0.01, when the epoch is more than 160, the learning rate is 0.001. In addition, to prevent overfitting, batchnormation\cite{ioffe2015batch} and dropout\cite{srivastava2014dropout} are applied in Vgg-19 and our proposed Vgg-19-post, the parameter of dropout is set to 0.5.

\textbf{Residual networks-32}: In computer vision, the “level” of features increases with increasing depth of the network. Studies have shown that the depth of the network is an important factor in achieving good results. However, gradient dispersion/explosion becomes an obstacle to training deep networks, resulting in failure to converge. To address gradient dispersion/explosion, residual network\cite{he2016deep} is proposed by kaiming he. During training, batch$\_$size is set to 64, epochs is 250, iterations is 782, weight$\_$decay is set to 0.0001. when the epoch is less than 80, the learning rate is 0.1, when the epoch is between 80 and 122, the learning rate is 0.01, when the epoch is more than 22, the learning rate is 0.001. In this experiment, total layers is 32.

\textbf{DenseNet}: DenseNet is a convolutional neural network with dense connections. In this network, there is a direct connection between any two layers. That is to say, the input of each layer of the network is the union of the output of all the previous layers, and the feature map learned by the layer is also directly transmitted to all layers behind it are used as input. The calculation of the network is not redundant due to the reduction of the calculation amount of each layer of the network and the reuse of features. It can be said that DenseNet has absorbed the most essential part of ResNet, and has done more innovative work on this, which further improves the network performance. During training, batch$\_$size is set to 64, epochs is 300, iterations is 782, weight$\_$decay is set to 0.0001. when the epoch is less than 150, the learning rate is 0.1, when the epoch is between 150 and 125, the learning rate is 0.01, when the epoch is more than 225, the learning rate is 0.001. In addition, to prevent prevent the network from becoming too wide, growth rate is set to 12 and depth is 100.

\begin{table}
\caption{image classification accuracy of LeNet, LeNet-post, VGGNet, VGGNet-post, Residual networks-32, Residual networks-32-post, DenseNet, DenseNet-post on cifar10 and cifar100}\label{table3}
\centering
\begin{tabular}{lll}
\hline
Methods & cifar10 & cifar100\\
\hline
HOG &46.92 &38.24\\
\textbf{HOG-post} & 45.31&38.23\\
LeNet & 76.23& 58.43\\
\textbf{LeNet-post} &\textbf{77.52}&\textbf{60.50}\\
VGGNet &91.53 &70.56\\
\textbf{VGGNet-post} & \textbf{92.00} & \textbf{71.35}\\
Residual networks-32 & 91.68& 71.13\\
\textbf{Residual networks-32-post} &\textbf{91.82}&\textbf{71.34}\\
DenseNet & 93.81&80.24\\
\textbf{DenseNet-post} & \textbf{93.90}&\textbf{80.44}\\
\hline
\end{tabular}
\end{table}

From the table \ref{table3}, our postprocessing algorithm always get the better results than original method on cifar10 and cifar100. In particular, on cifar100, LeNet-post improves the correct rate by 2 percent over LeNet, which indicates the postprocessing algorithm has great potential to enhance feature representation.
\subsection{Fine-Grained Categorization}
In recent years, Fine-Grained Categorization has been a very hot research topic in the fields of computer vision and pattern recognition. The goal is to make a more detailed subclassing of large, coarse-grained categories. Fine-grained images have more detailed class classifications, and the differences between classes are more subtle. Often, only small local differences can be used to distinguish different categories. Compared with object-level classification tasks such as face recognition, the intra-class differences of fine-grained images are much larger, and there are many uncertain factors such as attitude, illumination, occlusion, and background interference. Therefore, fine-grained image classification is a challenging research task. Here, we choose several well-represented fine-grained classification algorithms to verify our proposed feature post-processing algorithm. 
\subsection*{Caltech-UCSD Birds-200-2011}
Caltech-UCSD Birds 200 (CUB-200)\cite{wah2011caltech} is an image dataset with photos of 200 bird species (mostly North American), including 6,033 images, each image has a corresponding bounding box, rough segmentation and attributes. 
\subsection*{Stanford cars}
 Stanford Cars \cite{KrauseStarkDengFei-Fei_3DRR2013} contains 16,185 images of 196 classes of cars. The data is split into 8,144 training images and 8,041 testing images, where each class has been split roughly in a 50-50 split. Classes are typically at the level of Make, Model, Year, e.g. 2012 Tesla Model S or 2012 BMW M3 coupe.

\begin{figure}{}\centering
\includegraphics[width=1\columnwidth]{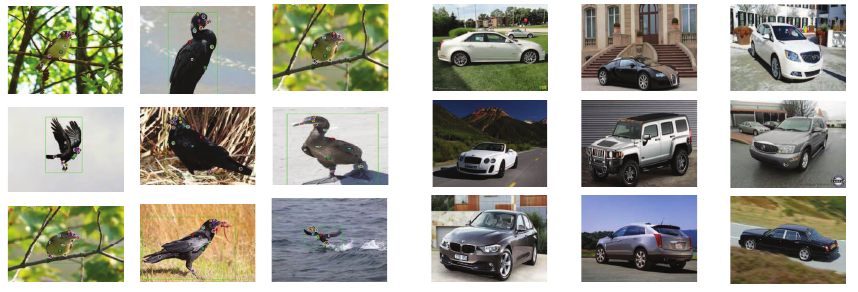} 
\caption{Examples from (left) birds dataset \cite{wah2011caltech}, and (right) cars dataset \cite{KrauseStarkDengFei-Fei_3DRR2013} employed in our experiments.} 
\end{figure}
\subsection*{Implementation Details}
As before, some classic fine-grained classification algorithms were used to verify the effectiveness of our proposed post-processing methods. 

\textbf{Bilinear CNN}:
Bilinear CNN\cite{lin2015bilinear} examines the interaction between different dimensions by computing the outer product of the convolution description vector. Since different dimensions of the description vector correspond to different channels of the convolution feature, and different channels extract different semantic features, the relationship between different semantic features of the input image can be simultaneously captured by the bilinear operation. During training, the images are resized to 224$\times$224 (the input size of the CNN). Vgg-16 \cite{simonyan2014very} is employed to be the M-Net and D-Net respectively and the parmeter of Vgg-16 is Pre-trained on imagenet\cite{deng2009imagenet}. The whole network is divided into two steps, first, we fix the parameters of the two VggNets and only learn the parameters of the fully connected layer, second, we remove the restrictions on the VggNets so that it can be trained, then, we can get the parameters of the final whole network.

\textbf{DFL-CNN}:
DFL-CNN \cite{wang2018learning} is an end-to-end way to learn the different mid$\_$level patches framework without the need for additional sections or border annotations. Then, our discerning patch does not need to be shared between classes and only need to be discernible. Therefore, our network is completely focused on classification, avoiding the trade-off between identification and positioning. For better results, we strictly follow the experiment on Caltech-UCSD Birds-200-2011 and Stanford cars. our DFL-CNN-post is only further processing on the extracted features of the original method.

\textbf{Cui's method}
Cui's method\cite{cui2018large} is a training scheme that achieves state-of-the-art results on large scale iNaturalist dataset by using higher resolution input image and fine-tuning to deal with long-tailed distribution. In addition, they further proposed a novel way of capturing domain similarity with Earth Mover’s Distance and shows better transfer learning performance can be achieved by fine-tuning from a more similar domain. During training, Subset B (585-class, Inception-v3) is used as our base feature extractor and other details you can refer to \cite{cui2018large}.

\textbf{MetaFGNet}
MetaFGNet\cite{zhang2018fine} is based on a novel regularized meta-learning objective with taking the target task into account during  pretraining, it  aims to guide the learning of network parameters so that they are optimal for adapting to the fine-grained visual categorization. In this work, we use 34-layer ResNet as the backbone of MetaFGNet (removed its last fully-connected). we strictly follow as the code\footnote{https://github.com/YBZh/MetaFGNet}, some other details you can find in it.

\begin{table}
\caption{Fine-Grained Categorization Accuracy of Bilinear CNN, Bilinear CNN-post, DFL-CNN, DFL-CNN-post, Cui's method, Cui's method-post, MetaFGNet, MetaFGNet-post on Caltech-UCSD Birds-200-2011 and Stanford cars}\label{table4}
\centering
\begin{tabular}{lll}
\hline
Methods & Caltech-UCSD Birds-200-2011 & Stanford cars\\
\hline
Bilinear CNN &84.1& 91.3\\
\textbf{Bilinear CNN-post} & \textbf{85.3}&\textbf{91.3} \\
DFL-CNN &86.7& 93.8 \\
\textbf{DFL-CNN-post} &\textbf{87.1} &\textbf{93.8}  \\
Cui's method &88.9 & 93.0\\
\textbf{Cui's method-post} & 88.6&\textbf{93.4} \\
MetaFGNet &87.1&91.2 \\
\textbf{MetaFGNet-post} &\textbf{88.0} &\textbf{91.8} \\
\hline
\end{tabular}
\end{table}
From the \ref{table4}, except the Cui's method on the Caltech-UCSD Birds-200-2011, our post-processing algorithms have achieved good results, which indicates that our post-processing algorithms are generally effective.


%


\section{conclusion}
In this paper, to address that image feature representations share a large common vector and a few top dominating directions, we propose a simple but effective postprocessing method to render off-the-shelf feature representations even stronger by eliminating the common mean vector and domainating directions from off-the-shelf feature representations.

\section*{Acknowledgment}

The authors would like to thank...

\ifCLASSOPTIONcaptionsoff
  \newpage
\fi



\bibliographystyle{IEEEtran}
\bibliography{IEEEtran}
\end{document}